\documentclass[10pt,twocolumn,letterpaper]{article}

\usepackage{3dv}
\usepackage{times}
\usepackage{epsfig}
\usepackage{graphicx}
\usepackage{amsmath}
\usepackage{amssymb}

\usepackage{bm}
\usepackage{subcaption}
\usepackage{booktabs}

\usepackage[normalem]{ulem}

\newcommand{\myparagraph}[1]{\vspace{0.2em}\noindent\textbf{#1}}

\usepackage{newfloat}
\DeclareFloatingEnvironment[
    fileext=los,
    listname={List of appendix figures},
    name=Figure,
    placement=tbhp,
    within=section,
]{apdxfig}

\usepackage[pagebackref=true,breaklinks=true,letterpaper=true,colorlinks,bookmarks=false]{hyperref}

\usepackage{caption}

\threedvfinalcopy 

\def\threedvPaperID{150} 

\ifthreedvfinal\fi
\begin{document}

\title{Grasping Field: Learning Implicit Representations for Human Grasps}



\newcommand*{\affaddr}[1]{#1}
\newcommand*{\affmark}[1][*]{\textsuperscript{#1}}
\newcommand*{\email}[1]{\small{\texttt{#1}}}
\author{
Korrawe Karunratanakul\affmark[1] \quad
Jinlong Yang\affmark[2] \quad
Yan Zhang\affmark[1]  \quad \\
Michael J. Black\affmark[2]  \quad 
Krikamol Muandet\affmark[2]  \quad
Siyu Tang\affmark[1]\\
\affaddr{\affmark[1]ETH Zurich} \quad \affaddr{\affmark[2]Max Planck Institute for Intelligent Systems}\\
\email{\{korrawe.karunratanakul,yan.zhang,siyu.tang\}@inf.ethz.ch} \quad \email{\{jyang,krikamol,black\}@tue.mpg.de}
}

\twocolumn[{%
\renewcommand\twocolumn[1][]{#1}%
\maketitle

\thispagestyle{empty}

\begin{center}
    \centering
    \includegraphics[width=\linewidth]{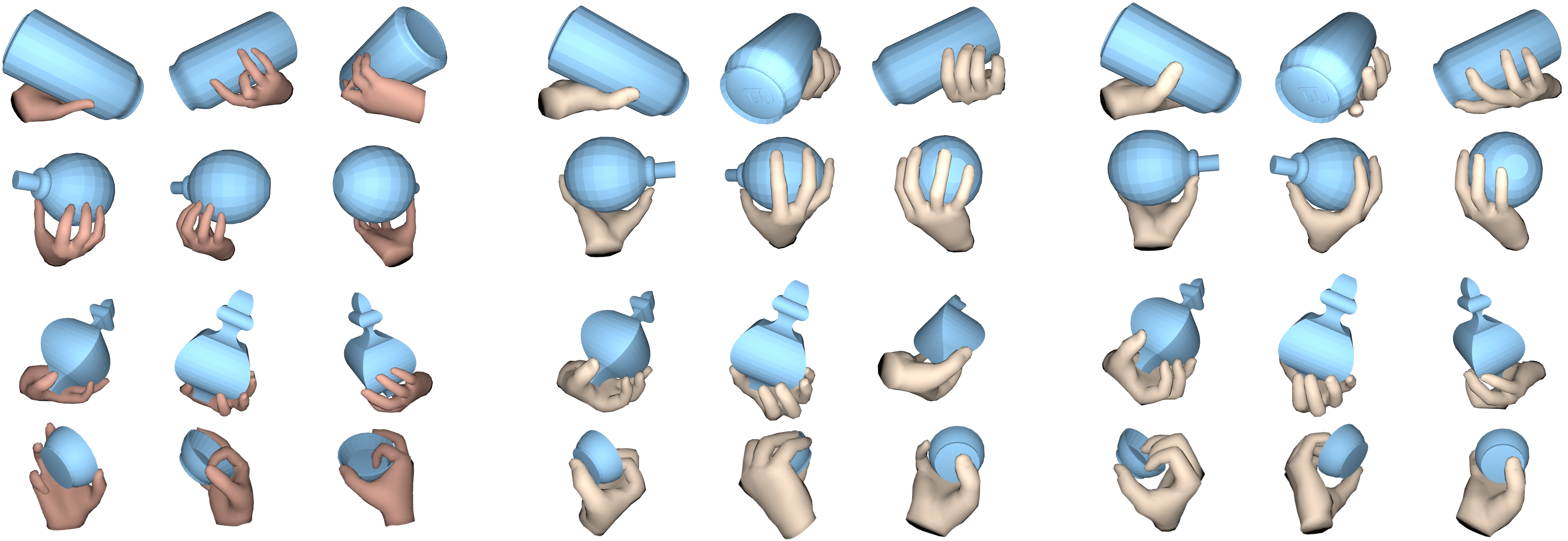}
    \captionof{figure}{Ground truth grasps and generated grasps. Each row corresponds to one object. Left three columns show the ground truth grasps, each from three different viewpoints. The middle three columns show one generated example, and the right three columns show another generated example. Note that these objects are never seen during training. See Appendix~\ref{app:qualitative} (Fig.~\ref{fig:sample_ho3d}) for more examples.}
    \label{fig:generative}
\end{center}%
}]

\begin{abstract}
Robotic grasping of house-hold objects has made remarkable progress in recent years.
Yet, human grasps are still difficult to synthesize realistically. 
There are several key reasons: (1) the human hand has many degrees of freedom (more than robotic manipulators); (2) the synthesized hand should conform to the surface of the object; and (3) it should interact with the object in a semantically and physically plausible manner. 
To make progress in this direction, we draw inspiration from the recent progress on learning-based implicit representations for 3D object reconstruction. 
Specifically, we propose an expressive representation for human grasp modelling that is efficient and easy to integrate with deep neural networks.
Our insight is that every point in a three-dimensional space can be characterized by the signed distances to the surface of the hand and the object, respectively.
Consequently, the hand, the object, and the contact area can be represented by implicit surfaces in a common space, in which the proximity between the hand and the object can be modelled explicitly. 
We name this 3D to 2D mapping as {\it Grasping Field}, parameterize it with a deep neural network, and learn it from data. 
We demonstrate that the proposed grasping field is an effective and expressive representation for human grasp generation.
Specifically, our generative model is able to synthesize high-quality human grasps, given only on a 3D object point cloud. 
The extensive experiments demonstrate that our generative model compares favorably with a strong baseline and approaches the level of natural human grasps. 
Furthermore, based on the grasping field representation, we propose a deep network for the challenging task of 3D hand-object interaction reconstruction from a single RGB image. Our method improves the physical plausibility of the hand-object contact reconstruction and achieves comparable performance for 3D hand reconstruction compared to state-of-the-art methods. Our model and code are available for research purpose at \url{https://github.com/korrawe/grasping_field}.

\end{abstract}


\section{Introduction}
\label{sec:intro}



Capturing and synthesizing hand-object interaction is essential for understanding human behaviours, and is key to a number of applications including augmented and virtual reality, robotics and human-computer interaction.  
Despite substantial progress, fully automatic synthesis of highly realistic human grasps remains an unsolved problem. The anatomical complexity of the human hand and the variety of manufactured and natural objects make it extremely challenging to pose the hand such that it interacts with the object in a natural and physically plausible way. Recent data-driven approaches explore deep learning technology to learn and leverage powerful object representations, yet they are mainly limited to simple robotic end effectors, such as parallel jaw grippers \cite{7139361}. In this work, we seek to understand: 1) what is an efficient and expressive representation for modeling hand-object interaction, that can facilitate realistic human grasp synthesis given an unseen 3D object; and 2) how can we learn such a representation from data.

Our key observation is that human grasping is rooted in physical hand-object {\em contact}. Through this contact, humans are able to grasp and manipulate objects naturally. To better model hand-object interaction, we must find a way to effectively represent the contact between hands and objects. 
To this end, we propose a novel interaction representation that is based on regressing a continuous function that we call the {\it Grasping Field}. The grasping field maps any 3D point to a 2D space, where each dimension of the 2D space indicates the signed distance to the surface of the hand and the object respectively (see Sec.~\ref{sec:gf_definition} for a formal definition).
Inspired by \cite{mescheder2019occupancy,OechsleICCV2019,park2019deepsdf}, we further utilize a deep neural network to parameterize the grasping field and learn it from data. 
As a result, the learned grasping field serves as a powerful representation to facilitate hand-object interaction modelling.

Based on the grasping field representation, we propose a generative model, in which we generate plausible hand grasps given an object point cloud.  
We show that our model can produce physically and semantically plausible synthetic grasps, which are similar to the ground truth. Generated grasps on unseen objects are shown in Fig.~\ref{fig:generative} and Fig.~\ref{fig:sample_ho3d}.

We further demonstrate the effectiveness of the grasping field representation by considering the task of 3D hand and object reconstruction from a single RGB image.
In recent work, Hasson et al.~\cite{hasson2019obman} introduce an end-to-end learnable model to reconstruct 3D meshes of the hand and object simultaneously, producing the state-of-the-art results on several datasets. Physical constraints, such as no interpenetration and proper contact, are enforced during the training.   
However, there are several drawbacks of their mesh-based representation for hand-object interaction modeling.
First, they heuristically pre-define regions of the hand that can be in contact with objects.
Second, their object representation is limited to objects of genus zero.
Third, the resolution of their contact inference is limited by the resolutions of the hand and object meshes. 
In contrast, with the grasping field representation, it is not necessary to first compute the hand and object meshes, and then compute the contact region. Instead, one can easily infer the contact region by querying the signed distances of input 3D points. Furthermore, the physical constraints, such as no inter-penetration and proper contact, can be efficiently computed and enforced. 
As demonstrated in our experiments, our model considerably reduces the interpenetration between the reconstructed hand and the object, and improves the quality of 3D hand reconstruction, compared with \cite{hasson2019obman}. 

In summary, our contributions are: (1) We propose the grasping field, a simple and effective representation for hand-object interaction; (2) Based on the grasping field, we present a generative model to yield semantically and physically plausible human grasps given a 3D object point cloud; (3) We further propose deep neutral networks to reconstruct the 3D hand and object given an RGB input in a single pass; (4) We perform extensive experiments to show that our method outperforms the baseline \cite{hasson2019obman} on 3D hand reconstruction and on synthesizing grasps that appear natural.



\section{Related work}
\label{sec:relatedwork}

\textbf{Human grasp and contact.} 
There is a large body of work on capturing and recognizing human grasps \cite{7298666,Glauser:Stretch-Glove:2019,4161001,Nakamura2017,pirk2017understanding,Yang_2015_CVPR}. 
Recently, \cite{Glauser:Stretch-Glove:2019} introduced a stretch-sensing soft glove to capture accurate hand pose without extra optical sensors. Puhlmann et al.~\cite{7759308} utilized a touch screen to facilitate the capturing process of human grasping. As physical contact is fundamental to hand-object interaction, 
researchers have proposed methods to capture and modeling contact from diverse modalities \cite{Brahmbhatt_2019_CVPR,2925927}, but these often interfere with natural movement.
Concurrent to this work, \cite{Brahmbhatt_2020_ECCV} proposed a new dataset of hand-object contact paired with RGB-D images.
Our work differs in that our focus is on learning an interaction representation, which is efficient and easy to interface with deep neural networks.

\textbf{Grasp synthesis.} 
Grasp synthesis is a longstanding problem in robotics and graphics, resulting in an extensive literature \cite{AntotsiouImitation,1492758,brahmbhatt2019contactgrasp,siggraph03Handrix,7139730,interactionCaptureSynthesis,4293017, Mahlereaau4984,Iasoniccv2011,Pollard2005, Siggraph91Hans,7138989,6225086,RealtimeHO_ECCV2016,10.1145/2508363.2508412}.
As early as 1991, Rijpkema and Girard \cite{Siggraph91Hans} proposed a knowledge-based approach to incorporate the role of the human hand, object, environment and animator for the task of computer-animated grasping. 
More recent works can be categorized into three types of approaches: analytic, data-driven and hybrid approaches. For the analytic approaches \cite{Krug2010,6225086}, the grasps are often synthesized by formulating the problem as a constrained optimization problem that satisfies a set of criteria measuring the stability or other properties of the grasps. The data-driven approaches \cite{PintoG16,7139361} often employ machine learning methods to learn representations for synthesizing grasps. An excellent survey of data-driven grasp generation is presented in \cite{10.1109/TRO.2013.2289018}. Recent hybrid approaches \cite{Mahlereaau4984,Mahler2017} combine analytic models and deep learning tools to synthesize grasps for various end effectors. 
Finally, the most related approaches to our work was presented in \cite{corona2020ganhand} and \cite{GRAB2020}, where neural networks are used to predict hand parameters of the MANO hand model \cite{MANO2017} given object information. In \cite{corona2020ganhand}, the model learns to predict the best grasp type from the grasp taxonomy \cite{feix2015grasp} according to the RGB images of the objects. Then, the predicted hands are optimized together with the object meshes to refine the contact points. While in \cite{GRAB2020}, the parameters are generated directly from the given Basis Point Set \cite{prokudin2019efficient} of the objects.
Our work differs from the previous works in that, by also considering the object distance field, we propose a learnable representation for modelling hand-object interaction that can be used without contact post-processing.
Empowered by deep neural networks, the learned representation enables us to synthesize realistic human hand grasping a given object naturally.

\textbf{Hand pose estimation.} 
Hand pose estimation is a long-standing problem, and various input modalities have been considered, e.g., RGB images \cite{baek2020weakly,Doosti_2020_CVPR,Iqbal_2018_ECCV,Mueller_2018_CVPR,8354158,Simon_2017_CVPR,spurr2018cvpr,zimmermann2017learning} or RGB-D and depth sensors \cite{5459282,kerkinECCV12,Malik_2020_CVPR,Mueller2017ICCV,Oberweger_2015_ICCV}.
Due to the lack of large scale 3D ground truth data, synthetic data has often been used for training \cite{Dibra2018a,hasson2019obman,hps2018}.
Recently, instead of estimating the hand skeleton, recovering the pose and the surface of the hand has become popular using statistical hand models, e.g., the MANO model \cite{MANO2017}, that can represent a variety of hand shapes and poses \cite{hampali2020honnotate,Zhang_2019_ICCV,Zimmermann_2019_ICCV}.
Using the template derived from MANO,~\cite{kulon2020weakly} show that it is also possible to regress hand meshes directly using mesh convolution.
In this work, we represent the 3D hand by a signed distance field, instead of a parametric hand model, due to the difficulty of incorporating object interaction into the model parameter space. For fair comparison with the parametric hand model representation, we fit the MANO model \cite{MANO2017} into our resultant signed distance fields. The experimental results indicate the advantage of our new interaction representation.

\textbf{Object model representation.} 
Learning 3D object models using various types of representations has also been explored \cite{Chen_2020_CVPR,groueix2018,kato2018renderer,li2016fpnn,Maturana2015iros,michalkiewicz2019deeplevel,PointSet2017,wang2018pixel2mesh,marrnet}.
Recently, the community has focused on using the implicit functions such as the Signed Distance Function (SDF)~\cite{park2019deepsdf}, Occupancy Networks~\cite{mescheder2019occupancy}, Implicit Field~\cite{chen2018implicit_decoder}, and their derivations~\cite{Genova_2020_CVPR,genova2019learning}, as these can model arbitrary object topology with adjustable resolution.
Due to these advantages, we also adopt implicit functions to capture hand-object interaction. 


\textbf{Hand-object interaction.} 
Reconstructing hand and object jointly has been studied with both RGB input and RGB-D input \cite{Pham_2018_TPAMI,Rogez:ECCVw:2014,Rogez:CVPR:2015,Romero09,Shan_2020_CVPR,Supancic:ICCV:2015,Tsoli2018,Tzionas:IJCV:2016,Tzionas:ICCV:2015,wang2019learning}.
Recently, Hasson~et al. \cite{hasson2020leveraging,hasson2019obman} achieved promising results on explicitly modeling the contact by combining a parametric hand model MANO \cite{MANO2017}, with the mesh based representation for the object. 
As data for hand-object interaction is limited, we opt to use their synthetic dataset, the ObMan dataset \cite{hasson2019obman}, which is sufficiently large for training a neural network. 
Our work differs from previous hand-object reconstruction work mainly by focusing on the novel representation of contact and learning both hand and object in the signed-distance space, which allows arbitrary shape modelling and easier distance field manipulation. Furthermore, we go beyond the reconstruction task by proposing generative models to synthesize realistic human grasps given a 3D object. 

\section{Method}
\label{sec:method}

\subsection{Grasping field}
\label{sec:gf_definition}

The \textit{grasping field} (GF) is based on the signed distance fields of the object and the hand, formally defined as a function $f_{\mathit{GF}}: \mathbb{R}^3 \to \mathbb{R}^2$, mapping a 3D point to the signed distances to the hand surface and the object surface, respectively. In this way, the contact and inter-penetration relations between the hand and the object can be explicitly and efficiently represented. 
Specifically, the hand-object contact manifold is given by $\mathcal{C} = \{\boldsymbol{x} \mid  f_{\mathit{GF}}(\boldsymbol{x}) = \boldsymbol{0} \text{ for } \boldsymbol{x} \in \mathbb{R}^3 \}$. 
The volume of hand-object inter-penetration is given by $\mathcal{I}=\{\boldsymbol{x} \mid f_{\mathit{GF}}(\boldsymbol{x})<\mathbf{0} \text{ for } \boldsymbol{x} \in \mathbb{R}^3 \}$.

Inspired by \cite{park2019deepsdf} and \cite{mescheder2019occupancy}, we propose to model $f_{GF}$ using a deep neural network, and learn it from data. 
Therefore, one can infer hand-object interaction in 3D space without the explicit hand and object surfaces. The learned GF can be considered as an interaction prior, which enables us to infer various grasping poses of the hand, only based on the 3D object.
Furthermore, in contrast to previous works, e.g. \cite{PROX:2019,hasson2019obman,PSI:2019}, which can only evaluate body-object interactions after obtaining the body and the object meshes, when using GF as the representation in hand-object reconstruction from images, we model the hand, the object, and the contact area by the implicit surfaces in a common space, largely improving the physical plausibility of the reconstruction.

According to the aforementioned merits of the GF, we use it to address two tasks in this paper; i.e.~hand grasp generation given 3D objects and hand-object reconstruction from RGB images. Different GF networks are designed specifically for different tasks.

\subsection{Grasping field for human grasp synthesis}
\label{sec:gf_pc}
In this section, we show how to use GF to synthesize human grasps. Given an object point cloud, the goal is to generate diverse hand grasps that interact with the object in a natural manner.

\begin{figure*}[t]
\centering
\subfloat[]{\includegraphics[width=0.65\linewidth]{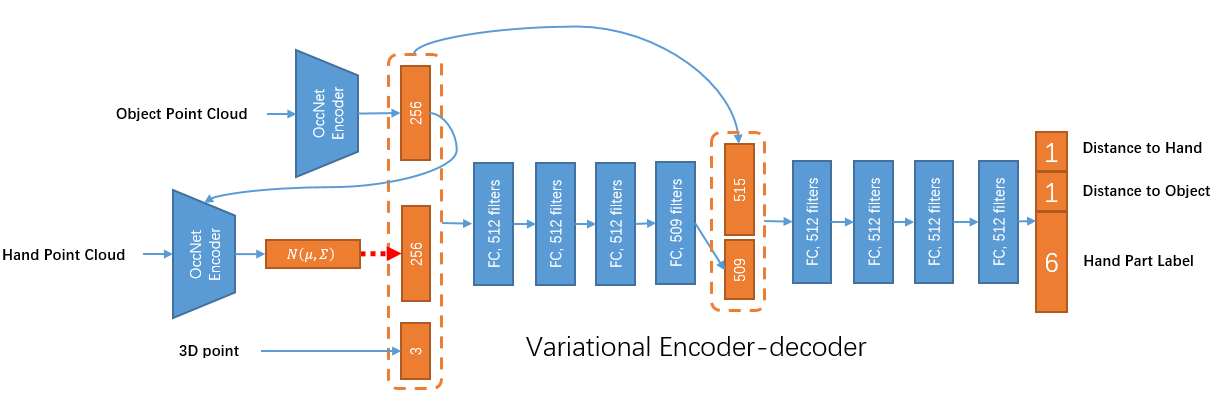}\label{fig:model_pcd}} 
\vspace{1em}
\subfloat[]{\includegraphics[width=0.25\linewidth]{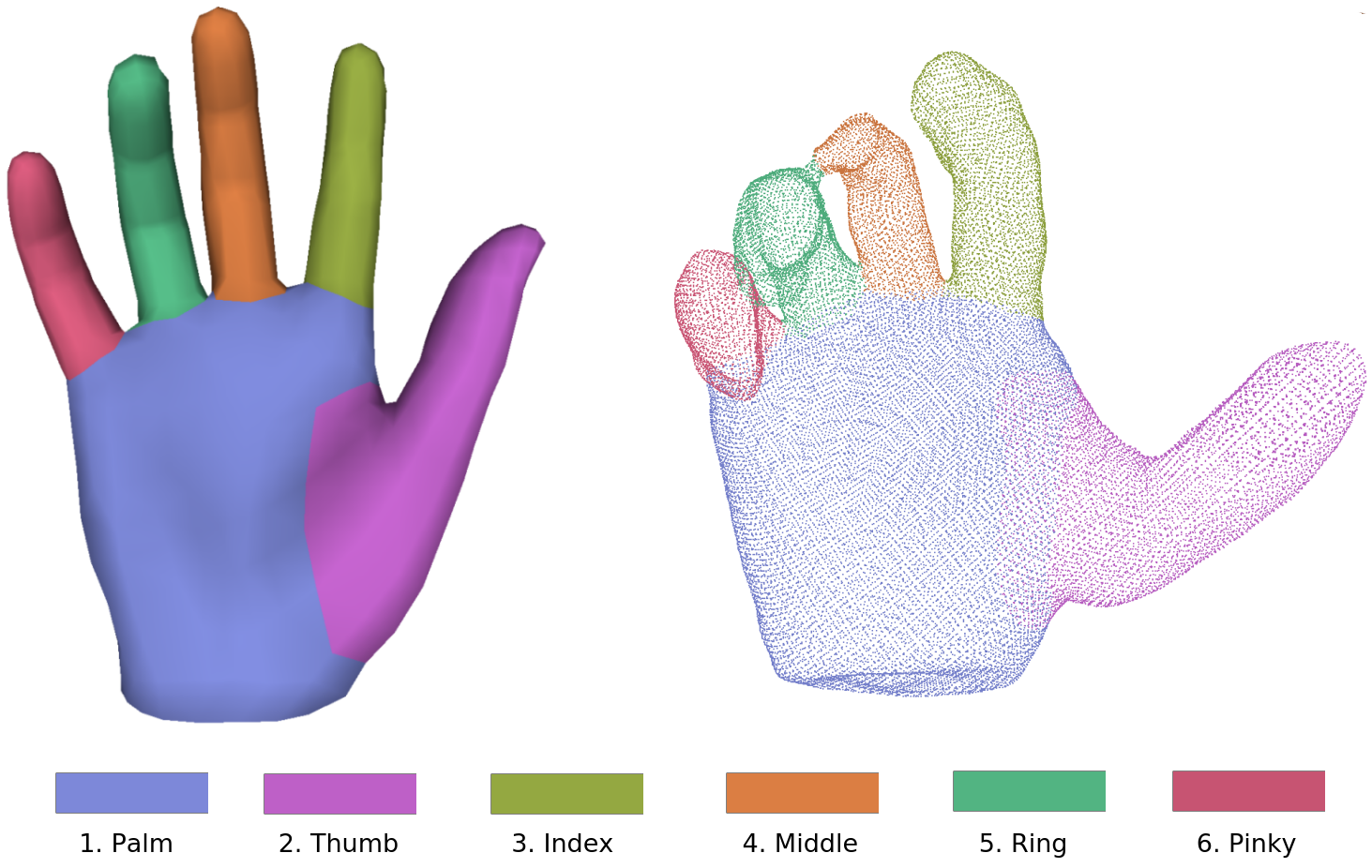}\label{fig:handLabel}}
\vspace{-1em}
  \caption{(a) Illustration of the generative grasping field network conditioned on an object point cloud. The red dashed arrow denotes sampling from a distribution. Network details are in Appendix~\ref{app:implemmentation}. (b) Illustration of hand part labels. Left is our hand part annotation on the MANO model. Right is an example of our \textit{predicted} surface points with hand part labels. }
\end{figure*}




\paragraph{Network architecture.} The network architecture is shown in Fig.~\ref{fig:model_pcd};
we adopt the encoder-decoder framework.
To extract features from point clouds, we use the PointNet encoder \cite{qi2017pointnet} with residual connection. The encoder is trained jointly with other network layers from scratch.
The encoder-decoder network takes a query 3D point, and two point clouds of the hand and the object as input, and produces the signed distances of the query point to the hand and the object surfaces. 
In addition, the encoded object point cloud feature is fed into the hand point cloud encoder, leading to a hand distribution conditioned on the object.
Note that this variational encoder-decoder network only requires both hand and object point clouds during the training. During inference, only the conditioning object point cloud and the query point are required. The hand features are sampled from the learned latent space, as in a standard VAE \cite{kingma2013auto}. 
The training loss consists of the following terms:

\myparagraph{The reconstruction loss $\mathcal{L}_{rec}$:} 
For each query point ${\bm x}$, the input object point cloud $p^o$ and the input hand point cloud $p^h$, the reconstruction loss is designed for the hand and object individually: 
\begin{multline}
    \mathcal{L}_{rec} = |c(f_{CGF}({\bm x}, p^{o}), \delta ) - c(\text{SDF}_{p^{o}}({\bm x}),\delta )| \\
    + |c(f_{CGF}({\bm x}, p^{h}), \delta ) - c(\text{SDF}_{p^{h}}({\bm x}),\delta )|,
\end{multline}
where $f_{CGF}$ is the grasping field network (Fig.~\ref{fig:model_pcd}).  $\text{SDF}_{p^{h}}(\cdot)$ and $\text{SDF}_{p^{o}}(\cdot)$ are the ground truth SDF for the hand and object, respectively. In addition,
${c}(s, \delta) := \min (\delta,{max}(-\delta,s))$ is a function to constrain the distance $s$ within $[-\delta, \delta ]$. $\delta$ is set to 1cm in all experiments. 


\myparagraph{KL-Divergence $\mathcal{L}_{kl}$:} In order to generate new hand grasps, we use a KL-divergence loss to regularize the distribution of hand latent vector $h$, obtained from the hand point cloud encoder $h=E_h(p^h | p^o)$, to be a normal distribution. The loss is given by 
\begin{equation}
    \label{eq:loss_kl}
    \mathcal{L}_{kl} = \textbf{KL-div} \left(N(\mu(h),\sigma(h)) || N(0,{\bm I})  \right),
\end{equation}
where $N(0,{\bm I})$ denotes a standard high-dimensional normal distribution, $\mu$ and $\sigma$ denotes mean and standard deviation. \
For generation, the hand latent vector $h$ is sampled from a standard normal distribution.

\myparagraph{Classification loss $\mathcal{L}_{cls}$:}
Besides predicting the signed distances of a query point, we also train the network to produce the hand part label of a query point to parse the hand semantically. To achieve this, we introduce a classification loss, which is given by a standard cross-entropy loss. The hand part annotation is based on the MANO model \cite{MANO2017} as illustrated in Fig.~\ref{fig:handLabel}. 

\subsection{Grasping field for 3D hand-object reconstruction from a single RGB image}
\label{sec:gf_image}

Our proposed grasping field is an expressive representation for modelling hand-object interactions in 3D. 
Here we address the challenging task of 3D hand-object reconstruction from a single RGB image, i.e.~$f_{CGF}: \mathbb{R}^3 \times \mathcal{I} \to \mathbb{R}^2$, in which $I \in \mathcal{I}$ is a 2D image. We model such a conditional GF by two types of deep neural networks and learn their parameters from data.

\begin{figure}[t]
\begin{center}
\centering
\subfloat[]{\includegraphics[width=0.96\linewidth]{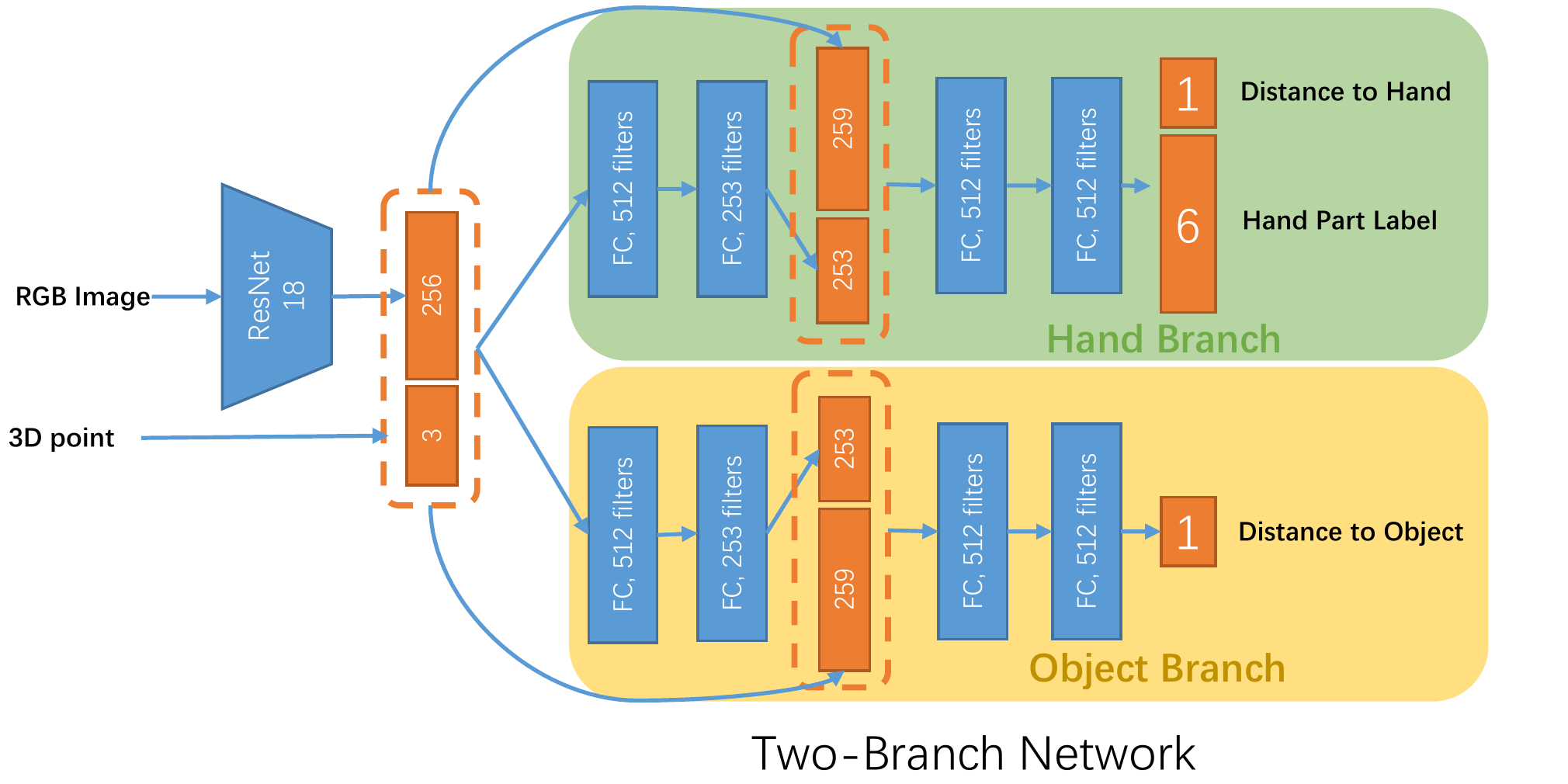}\label{fig:two_branch}} \\
\subfloat[]{\includegraphics[width=0.96\linewidth]{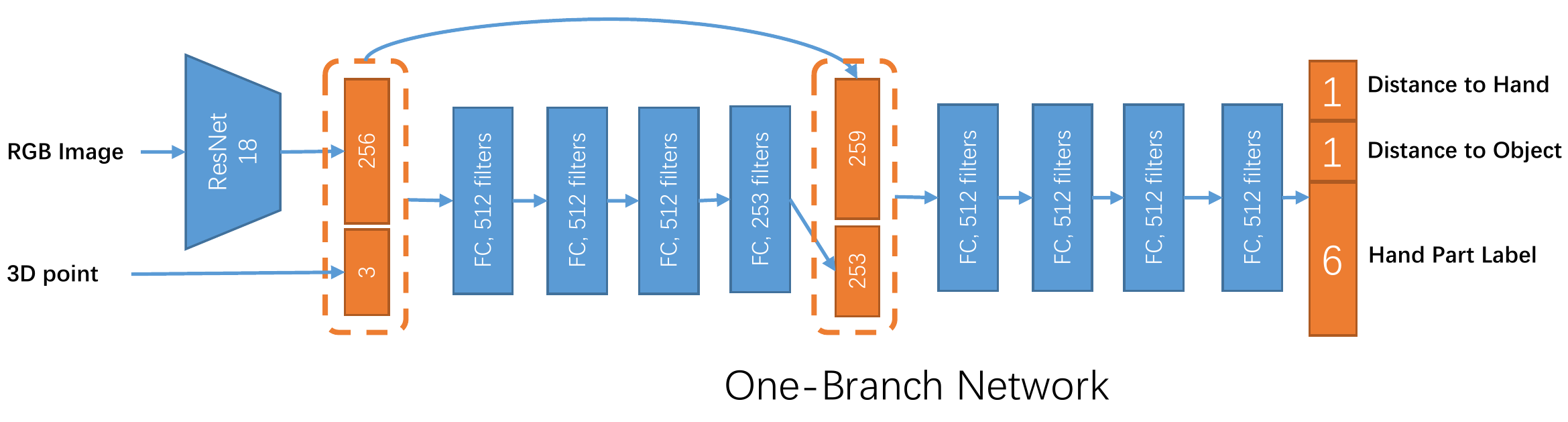}\label{fig:one_branch}}
\vspace{-0.5em}
  \caption{Two different network architectures of the GF conditioned on the image. The blue blocks denote network modules and layers. 
  A ReLU layer and a dropout layer (dropout ratio 0.2) are between every two consecutive fully-connected (FC) layers. The orange blocks denote feature vectors, and the feature dimensions are presented inside of these feature blocks. The orange dashed boxes denote feature vector concatenation. }
  \label{fig:model_image}
\end{center}
\end{figure}

\myparagraph{Network architecture.}
The network architectures are illustrated in Fig.~\ref{fig:model_image}, which are designed to recover both hand and object in a single pass. 
To enable a direct comparison with \cite{hasson2019obman}, the two-branch network is employed (Fig.~\ref{fig:two_branch}), which addresses hand and object individually. Similar to \cite{hasson2019obman}, we introduce contact and inter-penetration losses during the training to facilitate a better 3D reconstruction on the contact regions of the hand and the object.
To introduce hand-object interactions in early stages, we propose a one-branch network (Fig.~\ref{fig:one_branch}), which uses the same image encoder and has the same number of layers with the two-branch model. See Appendix~\ref{app:implemmentation} for architecture details.

\noindent The training loss consists of the following terms:

\myparagraph{The reconstruction loss $\mathcal{L}_{rec}$:} For each query point ${\bm x}$ and the input image $I$, the reconstruction loss is designed for the hand and object individually, and is given by
\begin{equation}
    \mathcal{L}_{rec} = \sum_{p \in \{ p^h, p^o \}} |c(f_{CGF}({\bm x}, I), \delta ) - c(\text{SDF}_p({\bm x}),\delta )|,
\end{equation}
in which $f_{CGF}$ is our conditinal grasping field network, and $\text{SDF}_p(\cdot)$ is the ground truth SDF for the component $p$ (hand or object). ${c}(s, \delta) := \min (\delta,{max}(-\delta,s))$ is the thresholding function to constrain the distance $s$ within $[-\delta, \delta ]$ as with the generative model proposed in Sec.~\ref{sec:gf_pc}.

\myparagraph{The inter-penetration loss $\mathcal{L}_{ip}$:}
To avoid surface inter-penetration between the reconstructed hand and object, we define the inter-penetration loss as
\begin{equation}
\mathcal{L}_{ip} = \sum_{{\bm x}} \max ( -\langle {\bm 1},  f_{CGF}({\bm x}, I) \rangle , 0),
\end{equation}
where ${\bm 1}$ is a 2D one-vector, and $\langle \cdot, \cdot \rangle$ denotes a dot product. This loss function actually penalizes the negative sum of predicted signed distances to the object and to the hand. If the hand and the object are separate and have no contact, the signed distance sum of every point in 3D space is always positive, and hence is ignored by our inter-penetration loss. On the other hand, if the hand and the object have inter-penetration, then this inter-penetration loss does not only penalize the points in the intersection volume, but also all 3D points in the space, indicating that the predicted hand and object are incorrect. 
Compared to the inter-penetration methods in \cite{PROX:2019,PSI:2019}, which only penalize the intersection volume, our loss applies stronger constraints.

\myparagraph{Contact loss $\mathcal{L}_{cont}$:}
Our proposed contact loss encourages hand-object contact, and is given by 
\begin{equation}
\mathcal{L}_C = \sum_{\bm x}{\min}(\alpha |f_{CGF}({\bm x}, I)|^2, 1),
\end{equation}
where $\alpha$ is a hyper-parameter. 
We can see that $f_{CGF}({\bm x}, I)=0$ corresponds to the hand-object contact surface. Therefore, it ignores points with predicted grasping field $ |f_{CGF}({\bm x}, I)|^2 \geq \frac{1}{\alpha}$, and only encourages points with $ |f_{CGF}({\bm x}, I)|^2 < \frac{1}{\alpha}$ to be the contact points. In our study, we empirically set $\alpha=0.005$ based on the hand-object interactions in the training data.
Finally, we employ the same \textbf{Classification loss $\mathcal{L}_{cls}$} as the one proposed in Sec.~\ref{sec:gf_pc}.


\subsection{From grasping field to mesh}
\label{subsec:fromGF2Mesh}
With the trained grasping field conditioned on images or point clouds, one can compute the signed distances to the hand and object of a query 3D point. To recover the hand, object and their interactions, we first randomly sample a large number of points, and evaluate their signed distances. The point clouds belonging to the hand and the object can be selected, according to point-object signed distances close to zero. Then, the hand mesh and the object mesh are obtain by marching cubes \cite{marchingcube}.

In addition, the hand mesh can be recovered by fitting the MANO  \cite{MANO2017} model to the hand point cloud. In this case, we can obtain hand segmentation, hand joint positions, and a compact representation of the hand simultaneously, according to the pre-defined topology in MANO. 

Denoting the MANO model by $\mathcal{M}(\beta, \theta)$ with the parameter $\beta$ and $\theta$ representing hand shape and pose respectively, we minimize 
$\sum_{l=1}^{6} d(\tilde{p}^h, \mathcal{M}(\beta, \theta)^l)$ 
to recover the hand configuration, in which $l$ denotes the 6 parts of hand, i.e. the palm and the 5 fingers, $\tilde{p}^h$ denotes the hand point cloud produced by our model, $\mathcal{M}(\beta, \theta)^l$ denotes the MANO hand mesh belonging to the hand part $l$, and $d(\cdot, \cdot)$ denotes the Chamfer distance \cite{fan2017point,park2019deepsdf}. The hand segmentation is shown in Fig. ~\ref{fig:handLabel}.

The implementation details are thoroughly presented in Appendix~\ref{app:implemmentation}.

\begin{figure*}[t!]
\centering
\includegraphics[width=1.0\linewidth]{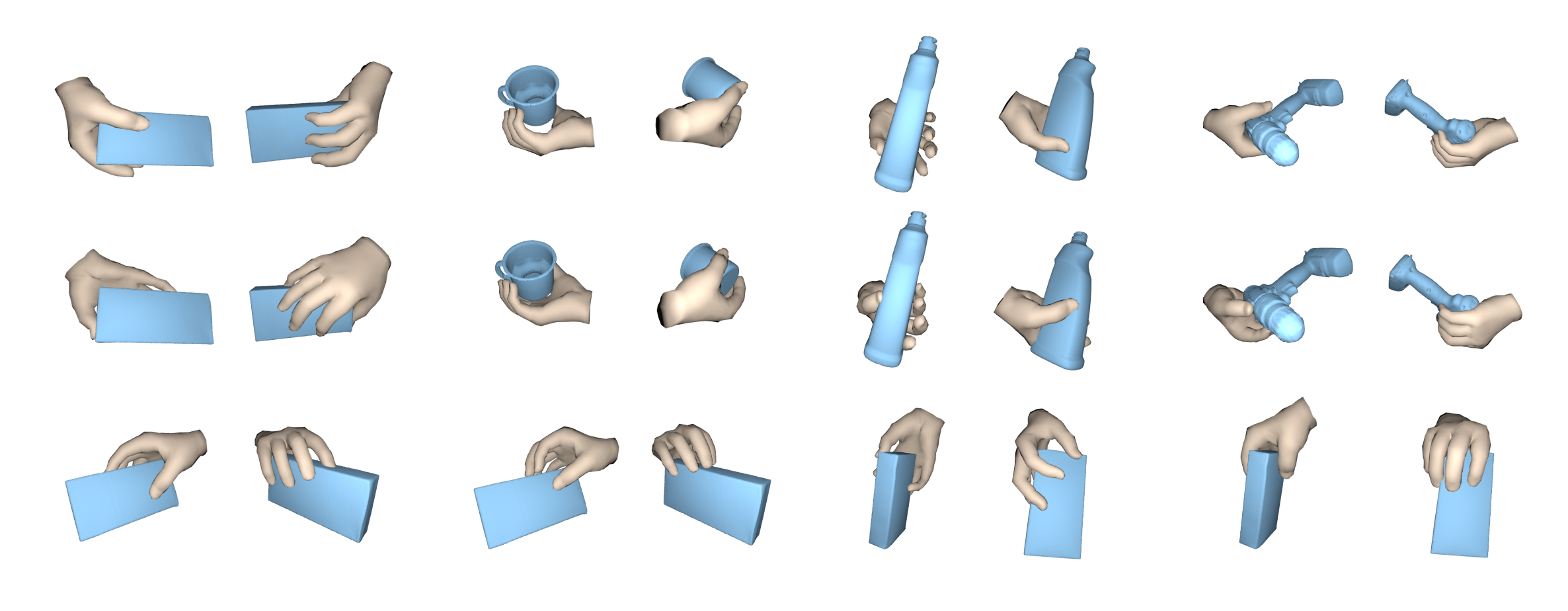}
  \caption{Generated grasps conditioned on objects from the HO3D dataset. Each pair shows the sampled grasp from two viewpoints. The model is trained only on the ObMan \cite{hasson2019obman} dataset.}
  \label{fig:sample_ho3d_main}
\end{figure*}

\section{Experiments}
\label{sec:experiments}

We demonstrate the effectiveness of the grasping field representation on two challenging tasks: human grasp generation given a 3D object and 3D hand-object reconstruction from a single image.  

\paragraph{Dataset.}
To train the generative model for human grasp synthesis, we need ground truth 3D meshes of interacting hands and objects. Unfortunately,  existing datasets often lack the desired properties.
The limitations include small dataset size and lack of 3D ground truth hand pose or shape. 
Consequently we use the synthetic \textbf{ObMan dataset}~\cite{hasson2019obman} to train our model. 
The data is generated from a statistical hand model, MANO \cite{MANO2017}, and 2772 object meshes covering 8 classes of everyday objects from the ShapeNet dataset~\cite{chang2015shapenet}.
Hand-object interaction is generated using a physics simulator, GraspIt~\cite{miller2004graspit}, resulting in high-quality hand-object interaction. 
Due to the limited number of grasp types in the \textbf{FHB dataset} \cite{garcia2018fhb} and the \textbf{HO-3D dataset} \cite{hampali2020honnotate}, they are not suitable for training the generative model (see Appendix~\ref{app_dataset_analysis}). 
Instead, we use them to test the generalization ability of the generative model trained on the \textbf{ObMan} grasps. 

For the 3D reconstruction task, we also mainly use the \textbf{ObMan dataset} for training and testing. To test the effectiveness of our network on real-world images, however, we follow the same approach as \cite{hasson20_handobjectconsist} to train and test on the \textbf{FHB dataset}. 

\paragraph{Evaluation metrics.}
Our goal is to generate physically plausible and semantically meaningful 3D human hand given an object. Therefore, 
we quantitatively evaluate the generated samples according to physics-based metrics and use large-scale perceptual studies to measure the visual realism of the grasps. For the 3D reconstruction task, we use Chamfer distance and hand joint error. Details of the evaluation metrics are in Appendix~\ref{app:eval_metrics}.

\textbf{(1) Physical metrics:} A valid human grasp implies stable hand-object contact without interpenetration. Consequently, we use the following evaluation metrics:
\textit{\textbf{a) Intersection volume and depth}}. The hand and object mesh are voxelized and the interpenetration depth is the maximum distance from all the intersected voxels to the surface of another mesh. \textit{\textbf{b) Ratio of samples with contact}}. We define a \textit{contact} between the object and the hand when any point on the surface of the hand is on or inside the surface of the object. We calculate the ratio of samples over the entire dataset that have interpenetration depth more than zero.
\textit{\textbf{c) Grasp stability}}. Using physics simulation \cite{coumans2013bullet}, we hold the hand constant, apply gravity, and measure the average displacement of the object's center of mass during a fixed time period. 

\textbf{(2) Semantic metric:} We perform perceptual studies using Amazon Mechanical Turk to evaluate the naturalness of our generated grasps. Details of the study are presented in Appendix~\ref{app:eval_metrics} and~\ref{app:user_study}.

\textbf{(3) 3D reconstruction quality:} We use the Chamfer distance between reconstructed and ground truth hand surfaces to evaluate the hand reconstruction quality as in~\cite{park2019deepsdf}.
Hand joint distance is computed following \cite{hasson2019obman,zimmermann2017learning}.

\begin{table*}[t]
    \centering
    \caption{Evaluation of the grasp synthesis on the objects from the ObMan test set, FHB and HO3D. GT* indicates that the ground truth grasps are obtained by fitting the MANO model to the data. Best results except the ground truth are shown in boldface. }
   \begin{tabular}{l c c c | c c c | c c c}
    \toprule
        &  \multicolumn{3}{c} {\bf ObMan}  &  \multicolumn{3}{c} {\bf FHB}    &  \multicolumn{3}{c}{\bf HO3D}\\
        & GT &  Baseline & GF   & GT*  &   Baseline & GF  & GT & Baseline & GF \\
        \midrule
        Contact ratio ($\%$) $\uparrow$      & - & 66.89 & \bf{89.4} &92.2 & 48.8& \bf{97.0}  & 93 &  44.3& \bf{90.1}\\
        Intersection vol. ($cm^3$) $\downarrow$ & - & 14.46 & \bf{6.05}  &16.6& \bf{9.65} & 21.9  & 10.5& \bf{5.86} &14.9\\
        Intersection depth ($cm$) $\downarrow$  & - & 0.94  & \bf{0.56}  & 1.99 & \bf{1.77} & 2.37 & 1.47 & \bf{1.01} &1.46\\
        Physics simulation ($cm$) $\downarrow$   &1.66  & 4.56  & \bf{2.07}  &6.69  & 8.59   & \bf{4.62} & 4.31  &8.25 & \bf{3.45} \\   
        & $\pm$ 2.42 & $\pm$ 4.57 & $\pm$ 2.81 & $\pm$ 5.48 & $\pm$ 3.67 & $\pm$ 4.48 & $\pm$ 4.42 & $\pm$ 4.18 & $\pm$ 3.92 \\
        Perceptual score $\{1...5\}$ $\uparrow$   & 3.24  & 2.40  & \bf{3.02}  & 3.49 & 2.43 & \bf{3.33} & 3.18 & 2.03 & \bf{3.29}\\
    \bottomrule
    \end{tabular}
    \label{tab:generation_model_full}
\end{table*}

\subsection{Evaluation: Human grasps generation}
\label{sec:eval:generation}

\myparagraph{Baseline.} To our knowledge, there is no previous model that learns to synthesize natural human grasps given a 3D object. Rather than randomly placing the hand around the object, we trained a strong baseline model for grasp generation. Specifically, we replace the decoder (i.e.~the grasping field) of our conditional VAE model (Fig.~\ref{fig:model_pcd}) with fully connected layers to regress MANO hand parameters. Then given a 3D object point cloud and a random sample, our baseline model generates MANO parameters directly. Generated grasps from the baseline are shown in Appendix~\ref{app:qualitative}. 

\myparagraph{Results.} We show the systematic quantitative evaluation of the generative method in Tab.~\ref{tab:generation_model_full} and qualitative results in Fig.~\ref{fig:generative},~\ref{fig:sample_ho3d_main} and Appendix~\ref{app:qualitative} (Fig.~\ref{fig:sample},~\ref{fig:sample_ho3d}). The baseline and the GF model are only trained on the \textbf{ObMan} training set, and tested extensively on the objects from the \textbf{ObMan} test set, \textbf{FHB} and \textbf{HO3D}. Our proposed GF performs substantially better than the baseline on \textbf{ObMan} and achieves comparable quality as the ground truth grasps. 
When the model is tested on the \textbf{FHB} objects, which are never seen during training, it achieves a comparable perceptual score compared to the ground truth grasps. 
Surprisingly, on \textbf{HO3D}, our synthesized grasps are judged more realistic than the ground truth grasps of real humans (3.29 vs 3.18).
These perceptual studies suggest that our method makes an important step towards the fully automatic synthesis of realistic human grasps. 

Regarding the physical plausibility, we observe that our model achieves a better contact ratio and grasp stability (physics simulation) than the ground truth grasps on \textbf{FHB} and \textbf{HO3D}. 
This is likely due to the GF results having a larger intersection volume.
One reason is that there are a very limited number of objects in these two datasets.
Some of the test objects are very different from the training objects, resulting in more inter-penetration for the generated grasps.
Overall, the combination of visual realism and grasp stability suggests that our results are approaching the level of natural human grasps.




\setlength{\tabcolsep}{4pt}
\begin{table*} [t!]
\caption{3D reconstruction results on \textbf{ObMan} (a) and \textbf{FHB} (b). 
2De refers to the 2 decoder model; 
L indicates the corresponding model is trained with the contact and interpenetration loss; GF-MANO refers to the MANO hand obtained by fitting the MANO model to the SDF.
}
\begin{minipage}{.7\linewidth}
\centering
\begin{tabular}{l|ll|l|ll|l|ll}
\toprule
Models  & \multicolumn{2}{l}{Hand error} & Joints & \multicolumn{2}{l}{Intersection} & Contact  & \multicolumn{2}{l}{Object Error} \\
 & Mean & Med  & (cm) & Vol & Depth & (\%)  & Mean & Med\\

\midrule
GF   & 0.419 & 0.283 & - & 0.65 & 0.32 & 90.8  & 12.8 & 6.4 \\
GF+L & 0.400 & 0.261 & - & 0.00 & 0.00 & 5.63  & 14.2 & 6.8\\
GF-2De  & 0.408 & 0.262 & - & 8.56 & 1.01 & 99.6 & 11.1 & 5.7\\
GF-2De+L& 0.384 & 0.237 & - & 0.23 & 0.20 & 69.6  & 11.7 & 5.9 \\

\midrule

GF-MANO  & 0.405 & 0.272 & 1.13 & 0.59 & 0.27 & 83.3  & - & - \\
GF-2De-MANO & 0.419 & 0.276 &  1.14 & 5.75 & 0.87 & 98.9 & - & - \\
\midrule
Hasson et al.~\cite{hasson2019obman}  & 0.533 & 0.415 & 1.13 & 6.25 & 1.20 & 94.8  & 6.7 & 3.6 \\
\bottomrule
\end{tabular}
\captionsetup{labelformat=empty}
\caption{(a)}
\end{minipage}
\begin{minipage}{.2\linewidth}
\begin{center}
\begin{tabular}{l c}
\toprule
Models & Hand Joints \\ 
 & (cm) \\
\midrule
GF-MANO\\(MANO joints) & 2.60\\
\midrule
GF-MANO \\(FHB marker) & 2.94\\
\midrule
Hasson et al.*~\cite{hasson20_handobjectconsist} & 2.74\\
\bottomrule
\end{tabular}
\captionsetup{labelformat=empty}
\caption{(b)}
\end{center}
\end{minipage}
\label{table:imageResults}
\end{table*}

\subsection{Evaluation: 3D hand-object reconstruction } \label{subsec:result_image_condition_v2}
Apart from serving as a powerful representation for the synthesis task, the proposed GF also facilitates the 3D reconstruction task.
In the following, we analyze the different network architectures and training losses proposed in Sec.~\ref{sec:gf_image}. We compare with the baseline method~\cite{hasson2019obman} on the \textbf{ObMan} dataset and  \cite{hasson20_handobjectconsist} on the \textbf{FHB} dataset. The results are summarized in Tab.~\ref{table:imageResults}. Due to many limiting factors of the real-world datasets such as \textbf{FHB} and \textbf{HO3D} (see Appendix~\ref{app_dataset_analysis} for detailed data analysis), learning a reasonable model for joint object and hand reconstruction is extremely challenging. Instead, to evaluate the effectiveness of the GF representation for 3D reconstruction on real-world images, we follow the setting of the latest work \cite{hasson20_handobjectconsist}, where the object 3D model is given as input. Note that this is a commonly used setting in previous works (e.g.~ \cite{hasson20_handobjectconsist,kehl2017ssd,Tekin_2019_CVPR}).

\myparagraph{Network design.}
We first analyze the two different network architectures presented in Fig.~\ref{fig:model_image} denoted with and without `2De' respectively. 
Both architectures achieve comparable performance for hand reconstruction, however differ significantly for intersection error, where the one decoder model achieves considerably better performance, due to the efficient joint modeling of hand-object interaction. Compared with the baseline \cite{hasson2019obman}, the intersection volume and depth are reduced from 6.25 and 1.20 to 0.65 and 0.32, respectively. The contact ratio are comparable among two architectures and baseline model.
All our models considerably improve the quality of hand reconstruction, compared with \cite{hasson2019obman}. The object reconstruction quality is behind hand quality for all model variations including the baseline model. 
Note that the \textbf{ObMan} dataset contains more than 1600 objects from 8 different classes. The object reconstruction performance is decreased as it remains unclear how to learn the implicit representations to reconstruct a large variety of object classes with a single model \cite{mescheder2019occupancy,park2019deepsdf} and such a task is beyond the focus of this work.  
Please see Appendix \ref{app:qualitative} (Fig.~\ref{fig:recon}) for visualization.


\myparagraph{Training losses.}
The effect of the contact and interpenetration loss ($+$L) is shown in Tab.~\ref{table:imageResults} (a), 
when the loss is imposed during the training of the two-decoder network, the intersection volume and depth are reduced and the overall quality of the interaction is considerably improved. In contrast, for the one-decoder model, our observation is that, for a large portion of 3D points, the signed distances to the object and to the hand are highly correlated, the model that jointly predicts both signed distance values does not need to enforce this auxiliary training loss.




\myparagraph{MANO fitting.}
As shown in Tab.~\ref{table:imageResults} (a), MANO fitting (indicated by GF-MANO) does not have a substantial influence on the reconstruction quality. This implies that on the one hand, the reconstructed hand of our GF model is realistic enough without a statistical model to regularize it, and that on the other hand, the output hand part labels are accurate enough for us to fit the MANO model and retrieve hand joints or shape parameters for applications that need these without undermining the shape and contact estimation.
A qualitative illustration is presented in Fig.~\ref{fig:mano_fit}.


\myparagraph{Hand reconstruction on real-world images.} 
To analyse the effectiveness of the proposed grasping field representation for the 3D reconstruction task on real-world images, we compare our method with the latest work \cite{hasson20_handobjectconsist} on the \textbf{FHB} dataset. Compared to \cite{hasson2019obman}, the key difference in~\cite{hasson20_handobjectconsist} is that the object is given as part of the input. We explore the same network architecture as \cite{hasson20_handobjectconsist} and only replace the decoder part with the grasping field.
The implementation details are presented in Appendix~\ref{app:implemmentation} (Fig.~\ref{fig:network_object_input}).

As stated in \cite{hasson20_handobjectconsist}, definition of hand joint locations vary between datasets. 
Without hand surface annotation in the FHB dataset,
it is difficult to train an accurate regressor that maps between the FHB markers and the MANO joints. Assuming that the joints are identically defined, we fit the MANO model to the FHB markers by minimizing the distance between the MANO joints and the FHB markers. Then the MANO joints obtained in such way are considered as our pseudo ground truth joints, and the obtained MANO surface  is used to supervise the training. 

We compare the predicted MANO joints with the pseudo ground truth joints as well as the original FHB markers assuming identical joints. 
As our model is not trained to optimise for the FHB marker locations, the reconstruction error is larger than~\cite{hasson20_handobjectconsist} as shown in Tab.~\ref{table:imageResults} (b). When we evaluate our prediction on the pseudo ground truth MANO joints, the reconstruction error decreases from $2.94 cm$ to $2.6 cm$. This suggests that the proposed grasping field representation is effective for the task of 3D hand reconstruction from a single image, achieving comparable performance with respect to the start-of-the-art.

\section{Conclusion and Discussion}
\label{sec:conclusion}

In this work, we propose a novel representation for hand-object interaction, namely the grasping field. 
Learning from data, the GF captures the critical interactions between hand and object by modeling the joint distribution of hand and object shape in a common framework.
To verify the effectiveness, we address two challenging tasks: human grasp generation given a 3D object and shape reconstruction given a single RGB image. The experiments show that the generated hand grasps appear natural and are physically plausible while the hand reconstruction achieves comparable performance as the state-of-the-art.

A limitation of our work is that there is no explicit modeling of the object functionality and human action in the current grasping field representation. In reality, a person holds an object differently based on different intentions. For instance, using a knife or passing it to someone else result in completely different human grasps. One promising future research direction is to incorporate human intention and object affordances into the grasping field for action specific grasps generation. Furthermore, 
we believe the proposed grasping field representation opens up avenues for several other future research directions. For instance, 3D human hand generation given only an object image and synthesizing the motion of hand-object interactions. 

\noindent \textbf{Acknowledgement.} We sincerely acknowledge Lars Mescheder and Michael Niemeyer for the detailed discussions on implicit function, Dimitrios Tzionas, Omid Taheri, and Yana Hasson for insightful discussions on hand interaction, Partha Ghosh and Qianli Ma for the help with VAE.
\textbf{Disclosure.} MJB has received research gift funds from Intel, Nvidia, Adobe, Facebook, and Amazon. While MJB is a part-time employee of Amazon, his research was performed solely at MPI. He is  an investor in Meshcapde GmbH.




{\small
\bibliographystyle{ieee}
\bibliography{main}
}

\clearpage

\begingroup
\onecolumn 

\appendix
\begin{center}
\Large{\bf Grasping Field: \\Learning Implicit Representations for Human Grasps\\ **Appendix**}
\end{center}

\setcounter{page}{1}

\section{Implementation Details}
\label{app:implemmentation}

In Sec.~\ref{sec:gf_pc} and Sec.~\ref{sec:gf_image}, we present the neural networks that are used for human grasps generation and reconstruction, respectively. Here we discuss the implementation details.


\subsection{Architecture}
In this section, we explain the network architectures used in our experiments.
The same decoder architecture is used in our image reconstruction and the hand generation tasks. 
We change the encoder architectures according to the input type. In our experiments, both the encoder and decoder are jointly trained end-to-end. 
Figure \ref{fig:decoder} illustrates the one-branch decoder with 8 fully-connected layers used in all tasks.

For image reconstruction, we use the ResNet18 \cite{he2016deep} model pretrained on the ImageNet dataset \cite{russakovsky2015imagenet} as an encoder. We change the last layer of the encoder to produce a latent vector of size 256 for the decoder.

For the point cloud input, we use two separated PointNet encoders \cite{fan2017point} with additional pooling and expansion layers presented in \cite{mescheder2019occupancy}.
In each encoder, 3D points are first mapped to 512-dimension feature vectors followed by 5 ResNet-blocks, producing a latent vector of size 256. 
The latent codes for hand and object are then concatenated to make a 512-dimension latent code.

For the hand generation task, we change the first layer in the hand encoder to produce a 256-dimension vector for each point then concatenate it with the 256-dimension object latent vector. Figure \ref{fig:vae_arch} shows the details of the point cloud model.

For image reconstruction with known objects, we assume that the object mesh in the normalized pose is given. We sample surface points from the given object and use a PointNet encoder to compute object latent vector of size 128. The object latent vector is then concatenated with a hand latent vector of size 128 from ResNet18 encoder, producing a latent code of size 256 for the decoder. The overview of the network is shown in Figure \ref{fig:network_object_input}.


\begin{apdxfig}[]
    \centering
    \includegraphics[width=\linewidth]{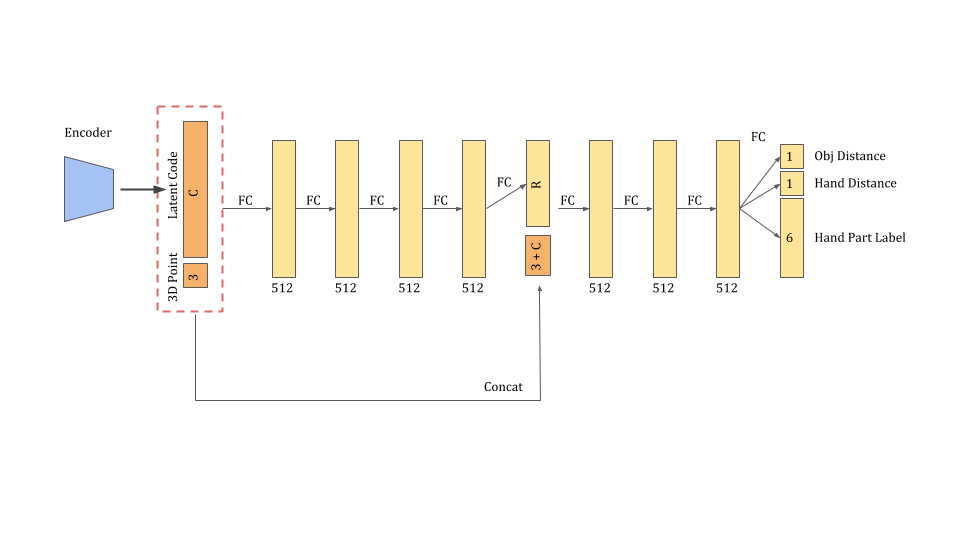}
    \caption{Decoder architecture. The fully-connected layers are denoted as ``FC'' in the diagram. The latent vector $C$ has 256 and 512 dimensions in the image reconstruction and conditional hand mesh generation task, respectively. The latent code from the encoder is concatenated with 3D point query then given to the decoder. The same latent vector is concatenated again at the middle of the decoder following \cite{park2019deepsdf}, with the concatenated vector $R$ having size 253 and 509, respectively. Every ``FC'' layer except the last layer is followed by a ReLu activation and a dropout layer with drop rate 0.2. The last layer produces the distance to object surface and the distance to hand surface along with hand part classification scores}
    \label{fig:decoder}
\end{apdxfig}

\begin{apdxfig}[]
    \centering
    \includegraphics[width=\linewidth]{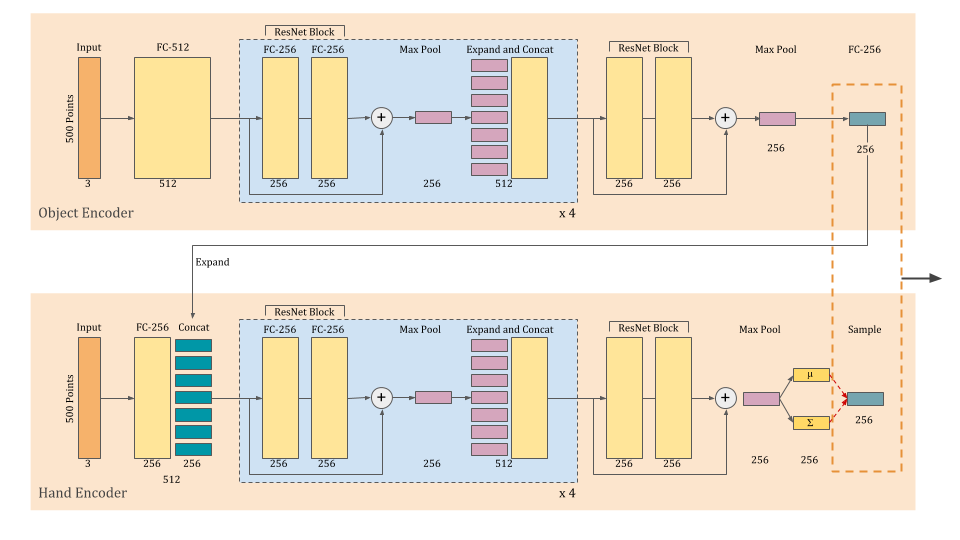}
    \caption{The encoder architecture for the conditional VAE. Each box represents a vector obtained from applying the layer written above. For the object encoder, we use the same architecture as the model for point cloud completion used in \cite{mescheder2019occupancy}. The hand encoder is conditioned on the latent code from the object decoder. The combined latent codes of hand and object are then concatenated with a 3D point and passed to the decoder}
    \label{fig:vae_arch}
\end{apdxfig}

\begin{apdxfig}[]
    \centering
    \includegraphics[width=\linewidth]{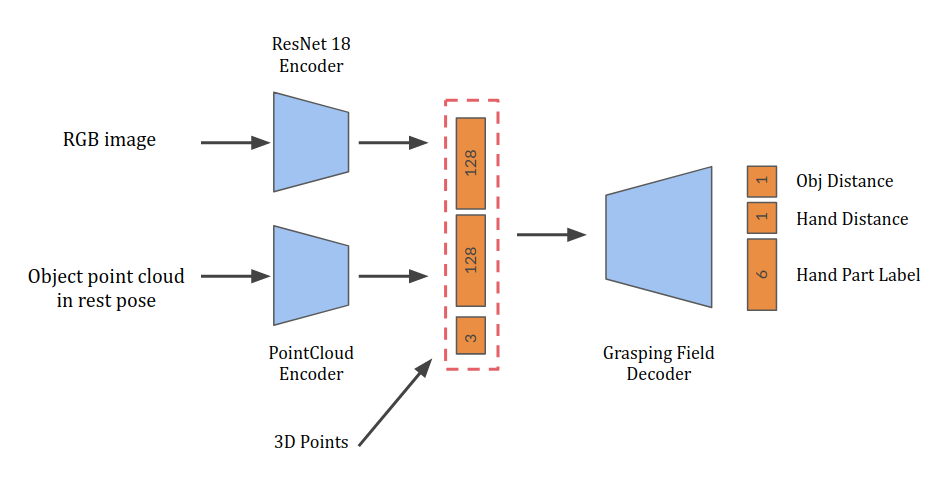}
    \caption{The architecture of the GF conditioned on the image, given a known object.}
    \label{fig:network_object_input}
\end{apdxfig}


\subsection{Data preparation}
\label{app_dataset_preparation}
To prepare the sampled 3D points and their distances to the hand and object surfaces for training,
we follow the point sampling method provided by \cite{park2019deepsdf}:
For each pair of the hand and object meshes, we translate both meshes such that the hand root joint is at the origin then scale them to fit in a unit cube. The scaling factor is the same for the entire dataset to ensure the hands are normalized across dataset.
After that, 40,000 points are sampled in a unit cube. Following \cite{park2019deepsdf}, 95\% of the total points were sampled near the surface to capture the details of both meshes.
For the Chamfer distance calculation, we sample 30,000 points from the surface of the ground truth mesh and reconstructed mesh following \cite{park2019deepsdf}. In case the reconstructed mesh contains more that one connected component, only the largest watertight connected component is retained.

\subsection{Training}

The contact loss is disabled in the beginning. When computing the reconstruction loss $\mathcal{L}_{rec}$, hand points to the object surface, and object points to the hand surface, are not considered until the contact loss is enabled. 
In our trials, we observe dramatic degradation when such a mask is not used or when the contact loss is enabled in the beginning.

For the generative GF network conditioned on an object point cloud, the KL loss, $\mathcal{L}_{kl}$, is employed in an annealing scheme; the loss weight is kept at 0 in the first 200 epochs and then linearly increased to 0.1 over the next 200 epochs. We find that such a annealing scheme is essential in our trials. Applying the KL loss in the beginning causes our generative network posterior to collapse.

In all experiments, we use Adam optimizer \cite{adam2014} with learning rate of $10^{-4}$ and decay it to $5\times10^{-5}$ at after 600 epochs. We train the models for 1,200 epochs without hand-part classification loss and another 100 epochs with the classification loss. Weight decay is used in all layers in the decoder. 

\subsection{Inference} \label{inference}
During inference, we use Marching Cube with resolution 128 to obtain hand and object meshes.
As the object can vary in size, we use a two-stage approach to dynamically scale the cube size in the Marching Cube algorithm. First, to find the boundary of the reconstructed meshes, we query equally space points in a unit cube centered at the origin point to locate the negative signed-distance values which indicate the inside of the mesh. 
Then, we query again with a cube that covers every negative-value point.
Using this approach, no mesh is produced if no negative point is found in the first stage.



\section{Dataset Analysis}
\label{app_dataset_analysis}
In this section, we provide detailed analyses on the \textbf{FHB} \cite{garcia2018fhb} and the \textbf{HO-3D} \cite{hampali2020honnotate}  datasets.
Although these datasets considerably contribute to the studies of hand-object interactions with detailed 3D annotation, our analyse shows that they might not be suitable for learning human grasps and modelling the accurate contact relation between hand and object. First, the number of objects and the types of grasps are limited. As shown in Tab.~\ref{table:datasetEval}, the number of object is 3  in the \textbf{FHB} dataset and 10 in the \textbf{HO3D} dataset. Second, the (pseudo) ground truth meshes of the interacting hand and object exhibit frequent interpenetration. 
Sampled ground truth meshes from the HO-3D dataset are illustrated in Fig.~\ref{fig:ho3d_inter}.

\setlength{\tabcolsep}{4pt}
\begin{table} [b!]
\caption{Characteristics of the FHB\textsubscript{c} and HO-3D dataset. The intersection volume and depth are calculated from the training sets and are considered when there is contact between hand mesh and object mesh.
For the FHB\textsubscript{c} dataset, the evaluation is done on the pseudo-ground truth meshes.} 
\begin{center}
\begin{tabular}{l|c|c|cc}
\hline
Dataset & \# of frames  & \# of objects & \multicolumn{2}{c}{Intersection} \\
 & (train/test) & & Vol(cm\textsuperscript{3}) & Depth(cm) \\
\hline

FHB\textsubscript{c} & 5082 / 5658 & 3 & 10.59 & 2.34 \\
\hline
HO-3D  & 66034 / 11524 & 10 & 10.91 & 1.56  \\
\hline
\end{tabular}
\end{center}
\label{table:datasetEval}
\end{table}
\setlength{\tabcolsep}{1.4pt}

\begin{apdxfig}[t!]
    \centering
    \includegraphics[width=\linewidth]{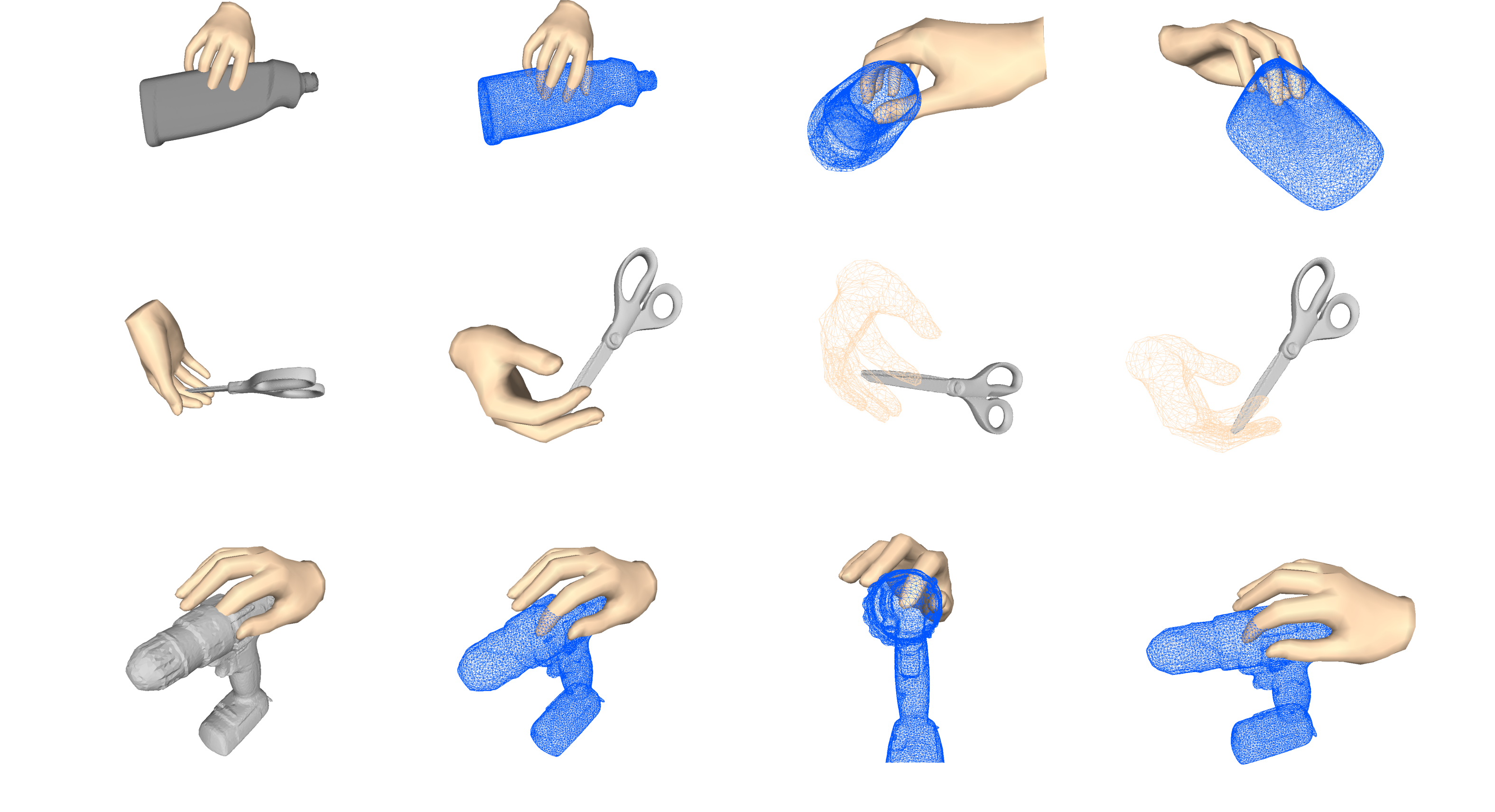}
    \caption{Ground truth meshes from the HO-3D dataset}
    \label{fig:ho3d_inter}
\end{apdxfig}

We evaluate the interpenetration between hand and object meshes quantitatively. We use the same evaluation metric as the one presented in the main paper, namely, the intersection volume (cm$^3$) and depth (cm) (Sec.~\ref{sec:experiments}).
The results are shown in Table \ref{table:datasetEval}.


For the HO-3D dataset, 91.94\% of the training examples exhibit hand-object contact. However, among these training examples, the average intersection volume and depth are 10.91 cm$^3$ and 1,56 cm respectively.


For the FHB dataset, 
we use the similar subset as the previous work \cite{hasson2019obman}, namely, we exclude the milk bottle related examples and the examples where the distance from hand joints to the object mesh is more than 1cm. We refer to this dataset as FHB\textsubscript{c}.  
We further fit the MANO hand model with the provided joint location. For FHB\textsubscript{c}, 97.1\% of the training examples have hand-object contact. However, similar level of intersection between hand and object meshes can be observed in Table.~\ref{table:datasetEval}.

Overall, the evaluation shows considerable intersection volume and depth of the training data. Therefore we use the ObMan dataset as our main training dataset, where the ground truth quality of the contact regions is more suitable for learning physically plausible human grasps.  

Furthermore, as shown in the experiment section (Tab.~\ref{tab:generation_model_full}), our generated grasps that are learned from the ObMan dataset obtain a higher perceptual score than the ground truth grasps from the HO3D data in the perceptual study, suggesting that the physical plausibility, i.e.~no interpenetration and proper contact, plays an important role on the naturalness of human grasps. 





% ...

\section{Details of the evaluation metrics}
\label{app:eval_metrics}

\myparagraph{Evaluation Metrics.}
For human grasps synthesis, our goal is to generate physically plausible and semantically meaningful 3D human hand given an object. Therefore, we propose to quantitatively evaluate the generated samples using physics metrics and a large-scale perceptual study to measure the perceptual fidelity. In addition, for the quantitative evaluation of our reconstruction networks, we use Chamfer distance and hand joint error.  

\textbf{(1) Physical metric:} A valid human grasp implies  hand-object contact without interpenetration. Naturally we propose with the following evluation metrics:

\textit{Intersection volume and \textit{depth}}. We follow \cite{hasson2019obman} to report intersection volume and depth. The hand and object mesh are voxelized using a voxel with edge length of 0.5cm. The interpenetration depth is the maximum distance from all points on the interpenetrated surface to another surface. If the meshes do not overlap, the interpenetration depth is defined as 0.

\textit{Ratio of samples with contact}. We define a \textit{contact} between object and hand when any point on the surface of hand is on or inside the surface of the object. To measure the performance of models on hand-object contact quality, we calculate the ratio of samples over the entire dataset that have interpenetration depth more than zero. As all of the samples in the dataset should have contact between hand and object, the best ratio of frames with contact is 100\%. 
A good hand and object reconstruction model should have high ratio of contact and small interpenetration volume and depth.

\textit{Simulation displacement}. Following \cite{hasson2019obman}, we use physics simulation to evaluate the stability of the grasps. In the simulated environment \cite{coumans2013bullet}, we fix the hand and measure the average displacement of the mass center of the object in a give time period. Small displacement suggests a stable grasp.

\textbf{(2) Semantic metric:} We perform perceptual studies on Amazon Mechanical Turk to evaluate the authenticity of our generated grasps. For each randomly generated sample, we render images from 6 different views, and request participants to score from 1 (low fidelity) to 5 (high fidelity).

\textbf{(3) 3D reconstruction quality:} We use the Chamfer distance between reconstructed and ground truth hand surfaces to evaluate the hand reconstruction quality. Surface distance is approximated by mean square point cloud Chamfer distance (cm$^2$) as implemented in~\cite{park2019deepsdf}. The MANO wrist is sealed to form a watertight mesh for fair comparison.
Joins distance is computed following \cite{hasson2019obman,zimmermann2017learning}.
After MANO parameters are recovered from the predicted hand mesh as described in Sec.~\ref{subsec:fromGF2Mesh}, we compute mean Euclidean distance over 21 joints following \cite{hasson2019obman,zimmermann2017learning}.
Note, since scale and global translation can not be determined by a single image, for each predicted hand, we optimize the scale and global translation to match the ground truth by minimizing the Chamfer distance between them. Similarly to hand, we also use Chamfer distance as measurement of object surface quality. The predicted object mesh is transformed according to the corresponding predicted hand transformation estimated from the above to align with the ground truth object mesh.

\section{Details of the perceptual study}
\label{app:user_study}
Figure \ref{fig:amt} shows the user interface for evaluating the generated grasps on the Amazon Mechanical Turk (AMT). Users are asked to rate the plausibility of the hand-object interactions individually. Each entry consists of images from six different views and is rated by three different users.

\begin{apdxfig}[]
    \centering
    \includegraphics[width=\linewidth]{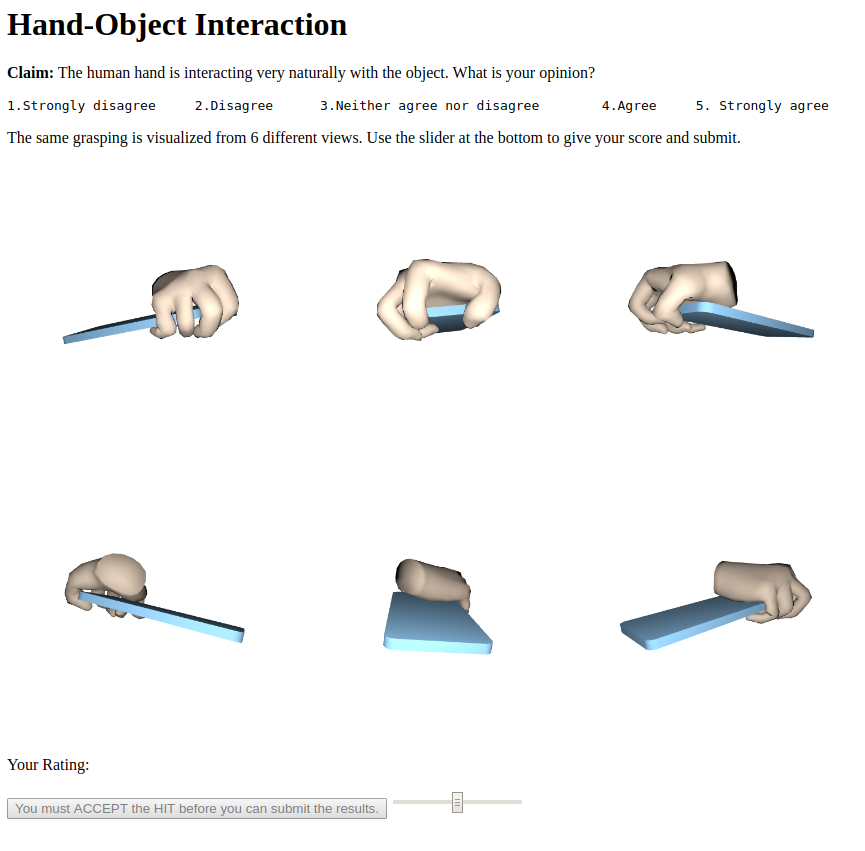}
    \caption{Amazon Mechanical Turk user interface for grasping evaluation}
    \label{fig:amt}
\end{apdxfig}


\section{Qualitative results}
\label{app:qualitative}

Figure \ref{fig:sample_baseline} shows the generated grasps from our baseline VAE model conditioned on the object surface point cloud. The MANO parameters are directly predicted by the decoder.

Figure \ref{fig:recon} shows the reconstruction results on the test images of the ObMan dataset \cite{hasson2019obman}. We observe that our model can recover hand meshes with proper interaction with the object. 

Figure \ref{fig:mano_fit} shows the comparison between reconstructed mesh before and after MANO fitting. The hand meshes also come from the single image reconstruction task. We observe that the MANO fitted meshes match the inferred meshes, even in the case where the rasterized hand mesh has merged fingers.

Figure \ref{fig:sample} shows \textbf{randomly sampled} grasps from our VAE model conditioned on the object surface point cloud. We observe that our model can generate a variety of grasps given an object. 
Figure \ref{fig:sample_ho3d} shows the generated results of the same model conditioned on the objects from HO3D datset. It should be note that this model is only trained on the ObMan dataset and have never seen these objects before.

\begin{apdxfig}[]
    \centering
    \includegraphics[width=\linewidth]{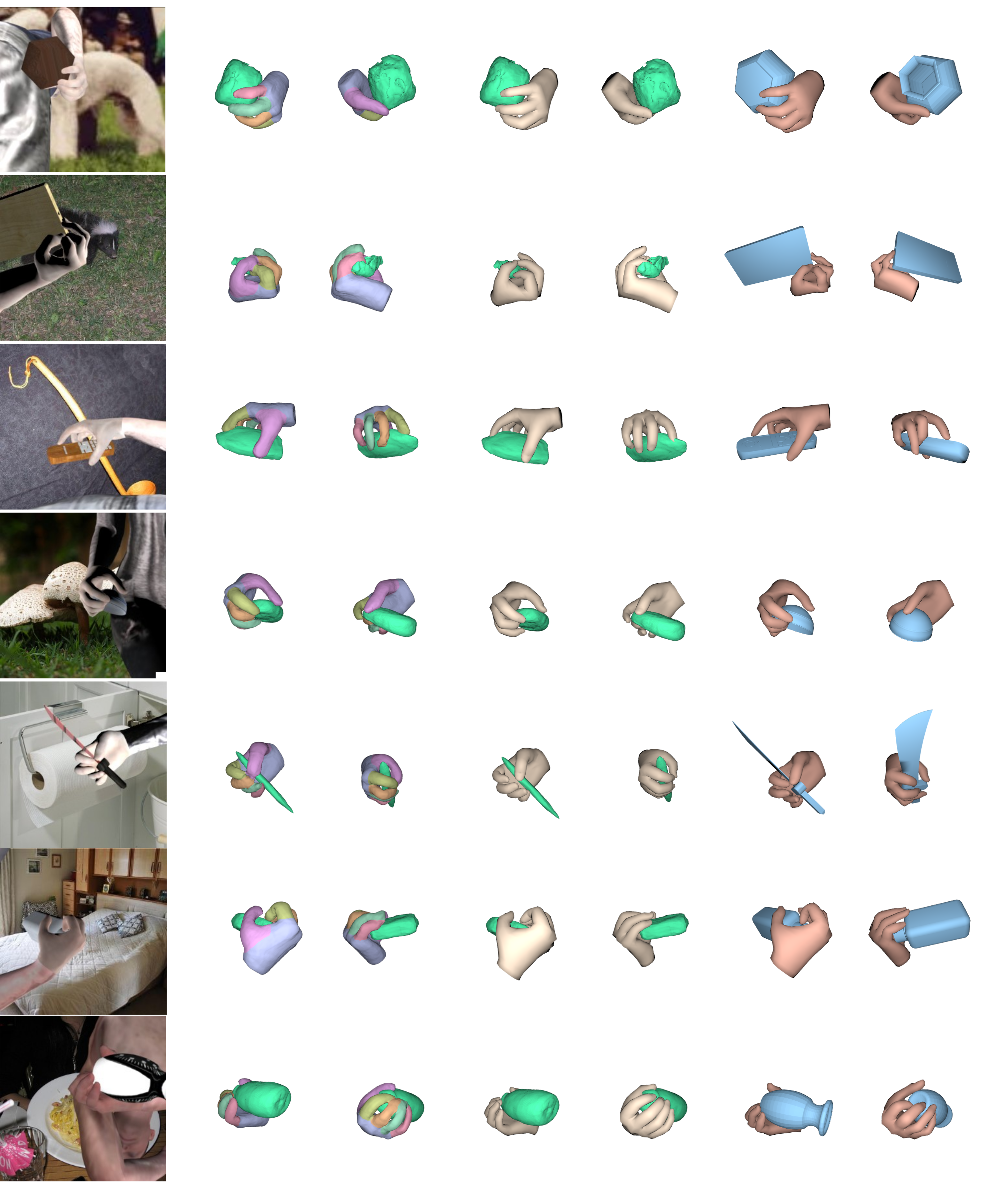}
    \caption{Reconstruction results from RGB images. From left to right: input images, recovered mesh from two different view points, MANO fitted hand prediction, ground truth hand and object.}
    \label{fig:recon}
\end{apdxfig}


\begin{apdxfig}[]
    \centering
    \includegraphics[width=\linewidth]{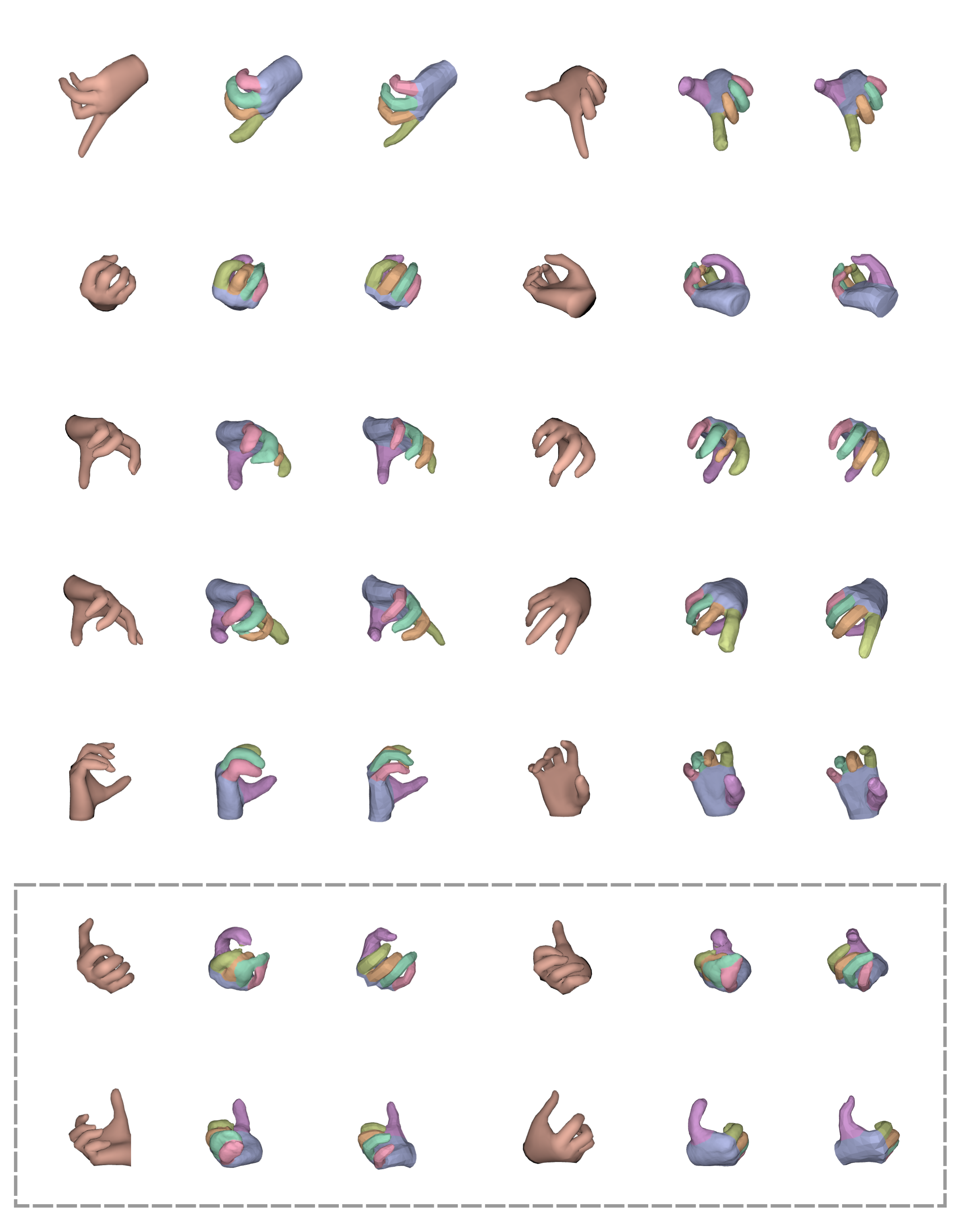}
    \caption{MANO fitting results on the reconstructions from RGB images. Each row shows a set of ground truth hand mesh, reconstructed hand, and MANO fitted prediction, from two different views. The last two rows demonstrate the robustness of our fitting method where the reconstructed meshes from the estimated SDF values are less satisfactory. 
    However, even with  merged fingers, we can still recover reasonable hand mesh.}
    \label{fig:mano_fit}
\end{apdxfig}


\begin{apdxfig}[]
    \centering
    \includegraphics[width=\linewidth]{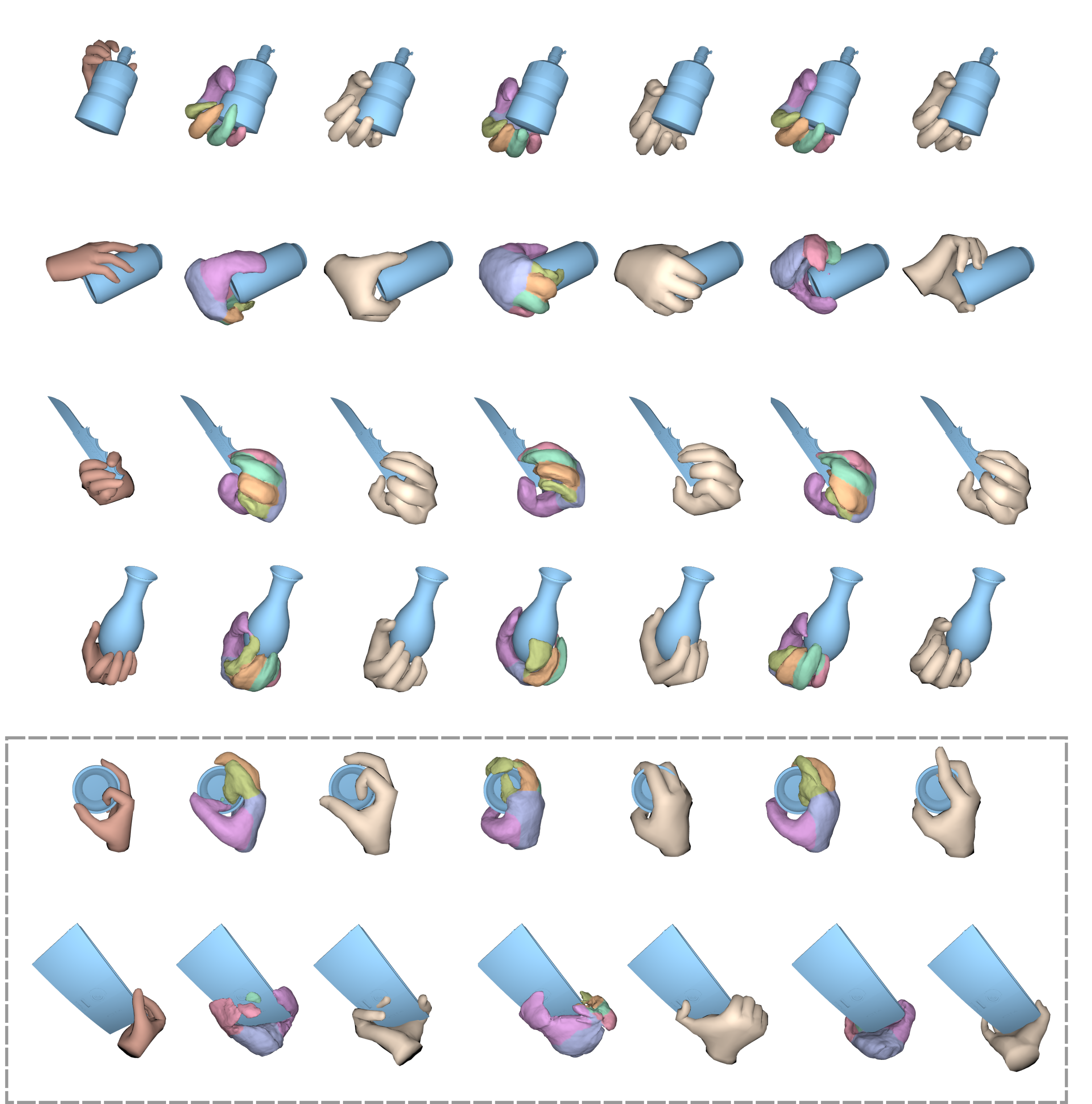}
    \caption{\textbf{Randomly selected} hands generated from the conditional VAE. Each row shows hand and object ground truth followed by three sets of sampled hand meshes, before and after MANO fitting, all from the same view. The samples presented on the two bottom rows are less satisfactory as the generated SDFs have artifacts and we can observe interpenetration between the fitted MANO hand meshes and the object meshes.}
    \label{fig:sample}
\end{apdxfig}

\begin{apdxfig}[]
    \centering
    \includegraphics[width=\linewidth]{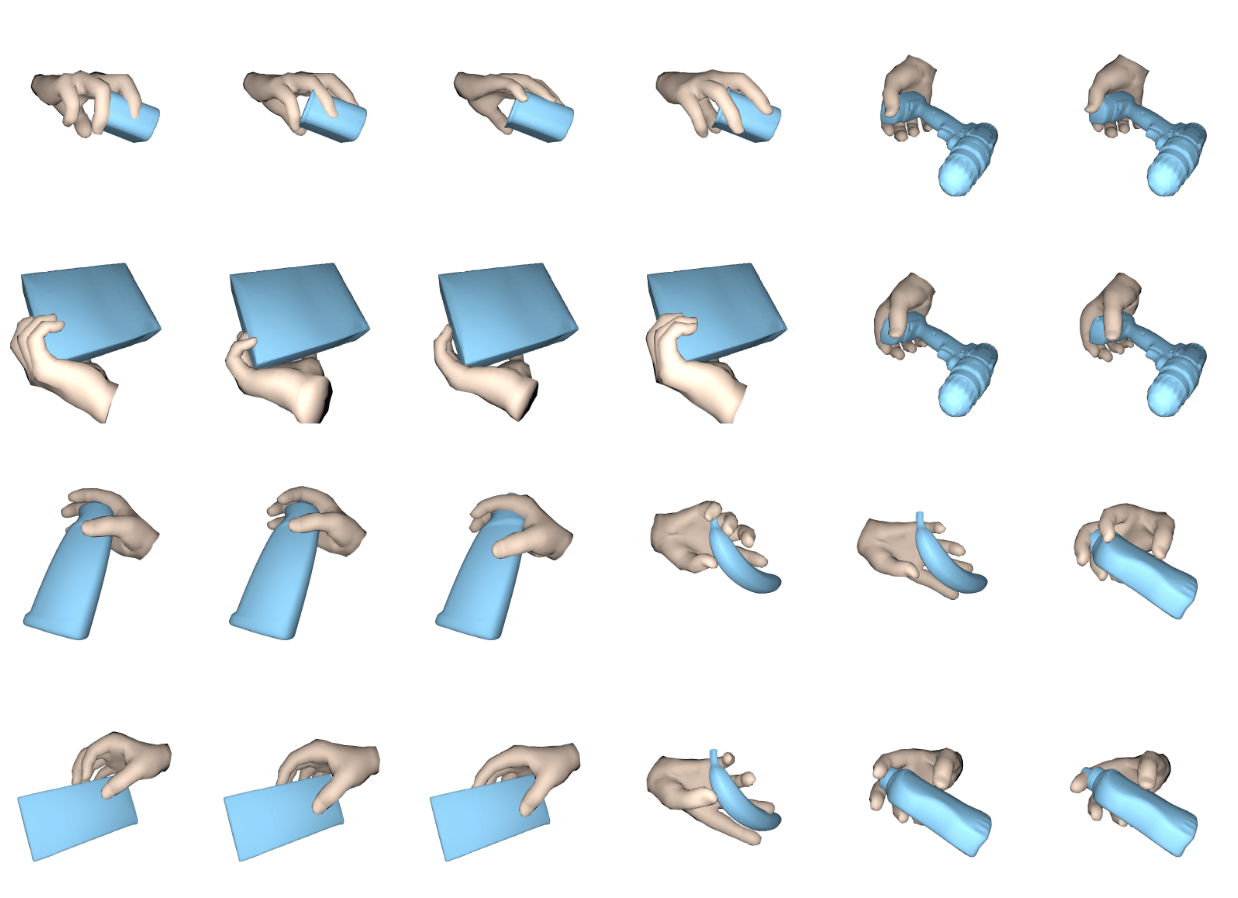}
    \caption{Hands generated from the VAE conditioned on objects from HO3D dataset. The model is trained only on the ObMan dataset.}
    \label{fig:sample_ho3d}
\end{apdxfig}

\begin{apdxfig}[]
    \centering
    \includegraphics[width=\linewidth]{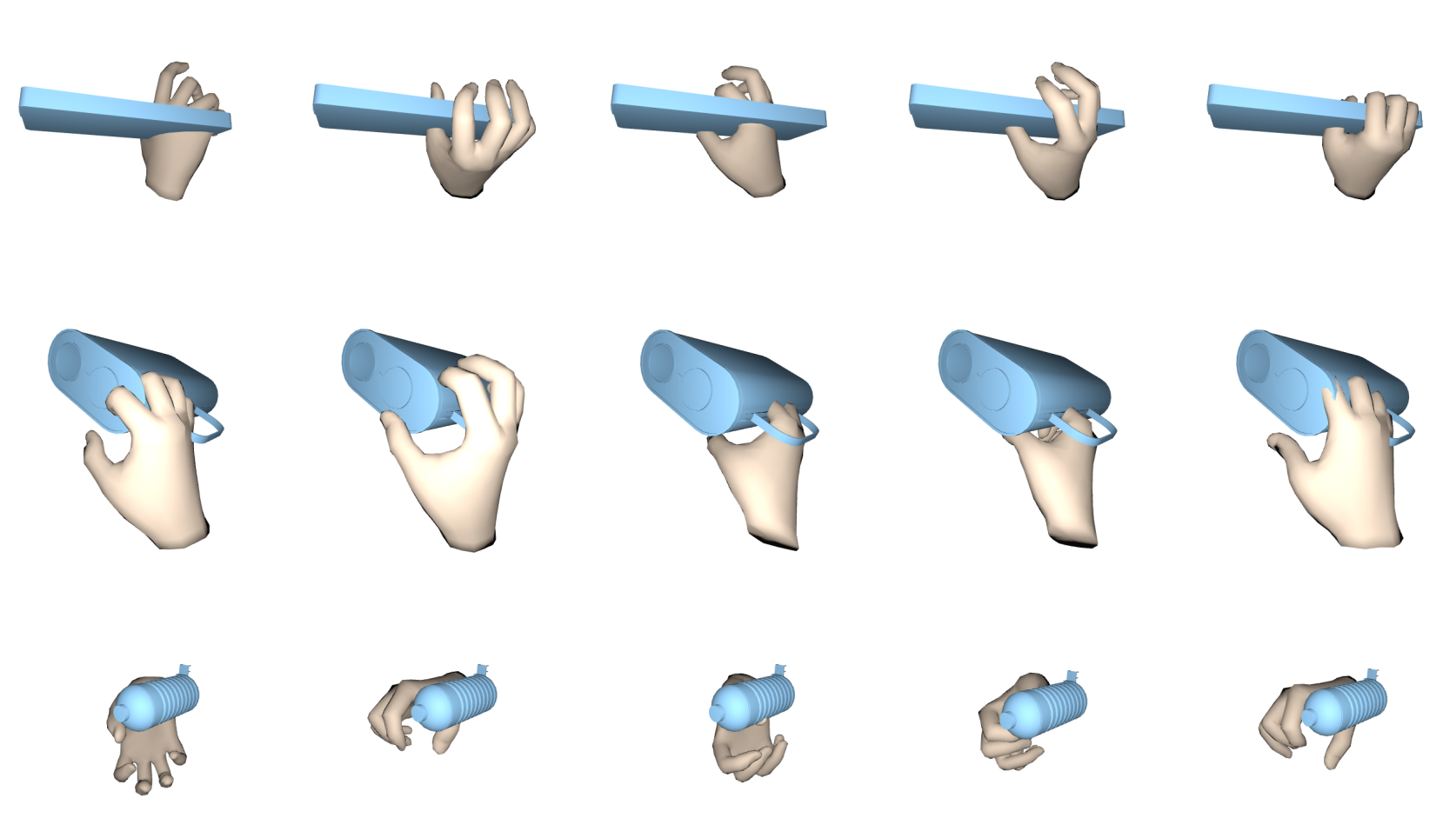}
    \caption{Each row shows five randomly sampled hands given an object from the baseline conditional VAE.  We can observe interpenetration between the hand meshes and the object meshes. In some cases, the hands are not in contact with the objects.}
    \label{fig:sample_baseline}
\end{apdxfig}

\begin{table*}[t]
    \centering
    \caption{Evaluation of the grasp synthesis on the objects from the ObMan test set, FHB and HO3D. GT* indicates that the ground truth grasps are obtained by fitting the MANO model to the data. Best results except the ground truth are shown in boldface. }
    \setlength{\tabcolsep}{8pt}
    \begin{tabular}{l c c | c c | c c }
    \toprule
        &  \multicolumn{2}{c} {\bf ObMan}  &  \multicolumn{2}{c} {\bf FHB}    &  \multicolumn{2}{c}{\bf HO3D}\\
        & GT &  GF   & GT*  & GF  & GT  & GF \\
        \midrule
        Contact ratio ($\%$) $\uparrow$      & - & {89.4} &92.2 & \bf{97.0}  & 93 & \bf{90.1}\\
        Intersection vol. ($cm^3$) $\downarrow$ & - &  {6.05}  & \bf{16.6} & 21.9  & \bf{10.5}  & 14.9\\
        Intersection depth ($cm$) $\downarrow$  & - & {0.56}  & \bf{1.99} & 2.37 & 1.47 & \bf{1.46} \\
        Physics simulation ($cm$) $\downarrow$   &1.66  & {2.07}  &6.69  & \bf{4.62} & 4.31 & \bf{3.45} \\
        Perceptual score $\{1...5\}$ $\uparrow$   & 3.24 & {3.02}  & 3.49 & \bf{3.33} & 3.18 & \bf{3.29}\\
    \bottomrule
    \end{tabular}
    \label{tab:ss}
\end{table*}

\end{document}